\documentclass[letterpaper, 10 pt, conference]{ieeeconf}  

\usepackage{amsmath,graphicx}
\usepackage{multirow}
\usepackage{amsmath}
\usepackage{amssymb}
\usepackage{cite}
\usepackage{colortbl}
\usepackage{xcolor}
\usepackage[pagebackref=true,breaklinks=true,colorlinks,bookmarks=false]{hyperref}

\DeclareMathOperator*{\argmax}{arg\,max}
\DeclareMathOperator*{\CE}{CE}

\usepackage{color}

\IEEEoverridecommandlockouts                              

\title{\LARGE \bf
SpiderMesh: Spatial-aware Demand-guided Recursive Meshing \\for RGB-T Semantic Segmentation
}

\author{Siqi Fan$^{1}$  Zhe Wang$^{1}$  Yan Wang$^{1*}$ Jingjing Liu$^{1*}$
\thanks{$^{1}$ Institute for AI Industry Research (AIR), Tsinghua University, {\tt\small \{fansiqi, wangyan\}@air.tsinghua.edu.cn};}
}

\begin{document}

\maketitle
\thispagestyle{empty}
\pagestyle{empty}

\begin{abstract}

For semantic segmentation in urban scene understanding, RGB cameras alone often fail to capture a clear holistic topology in challenging lighting conditions. Thermal signal is an informative additional channel that can bring to light the contour and fine-grained texture of blurred regions in low-quality RGB image. Aiming at practical RGB-T (thermal) segmentation, we systematically propose a Spatial-aware Demand-guided Recursive Meshing (SpiderMesh) framework that: $1)$ proactively compensates inadequate contextual semantics in optically-impaired regions via a demand-guided target masking algorithm; $2)$ refines multimodal semantic features with recursive meshing to improve pixel-level semantic analysis performance. We further introduce an asymmetric data augmentation technique M-CutOut, and enable semi-supervised learning to fully utilize RGB-T labels only sparsely available in practical use. Extensive experiments on MFNet and PST900 datasets demonstrate that SpiderMesh achieves state-of-the-art performance on standard RGB-T segmentation benchmarks.

\end{abstract}

\section{Introduction}

\begin{figure}[t]
  \centering
  \includegraphics[scale=0.46]{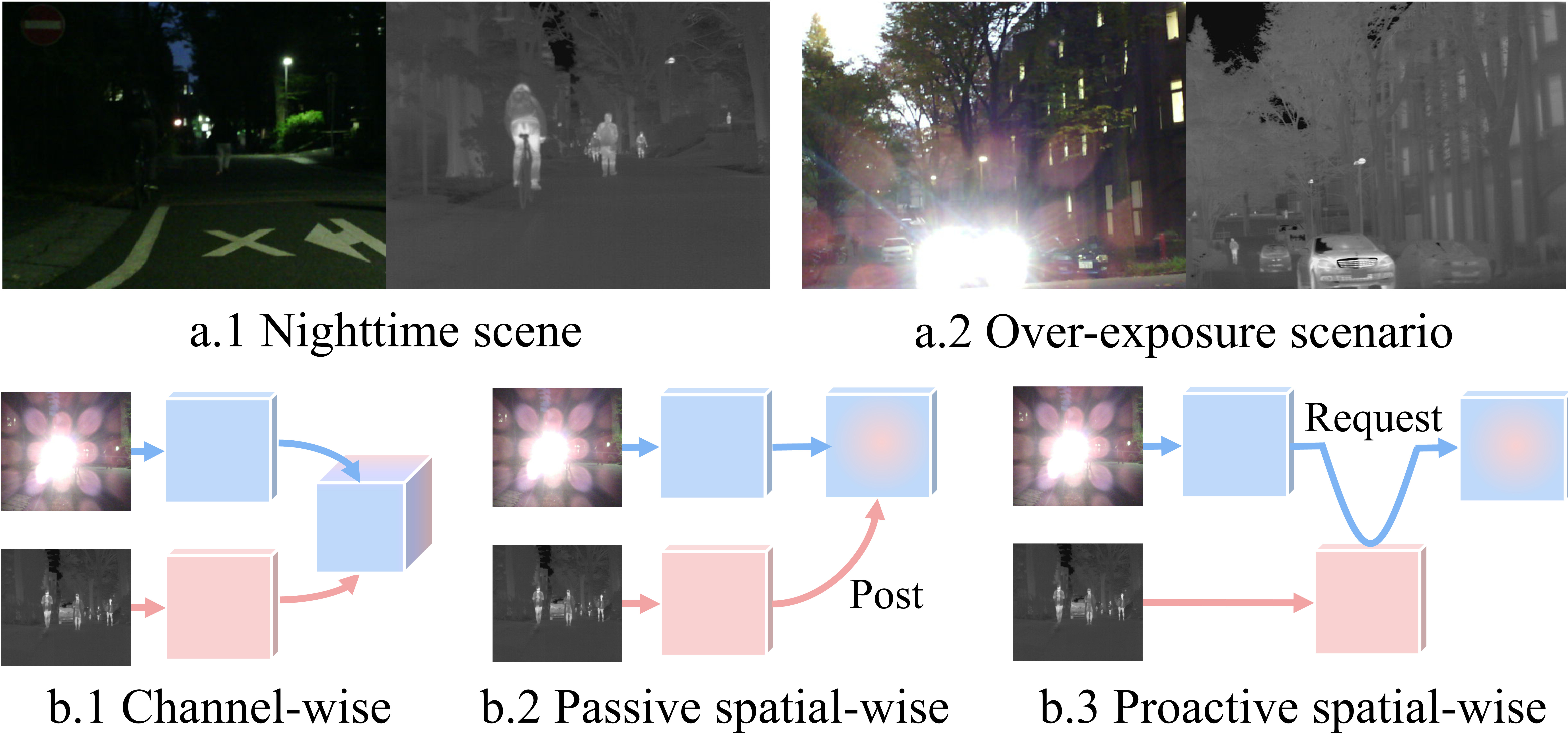}
  \caption{Some RGB regions are blacked out or blurred in nighttime or over-exposure scenes, while the corresponding thermal images are robust to varying illumination conditions (\textit{a}, images from MFNet dataset \cite{mfnet}). Compared with channel-wise and passive spatial-wise fusion,  proactive spatial-wise fusion can fully exploit extra semantic feature for targeted region (\textit{b}). Code is available at \href{https://github.com/leofansq/SpiderMesh}{GitHub}.}
  \label{fig:intro}
\end{figure}

To realize robust pixel-wise scene understanding in real-world urban environment, RGB-based semantic segmentation is often inadequate due to low image quality from poor lighting conditions, such as nighttime scenes and over-exposure scenarios (as illustrated in Figure~\ref{fig:intro} a). 

One effective way to overcome this is to dynamically adjust the exposure time of cameras. However, this introduces new challenges such as motion blur under long exposure time. To amend this, RGB-T brings in thermal sensors that are relatively robust to variation in illumination conditions \cite{mfnet, rtfnet, fuseseg, feanet, gmnet, mffenet, ABMDRNet, mitigating, 10103760, li2022rgb, 9987529, zhou2022mtanet, Zhang_2023_CVPR}. Unlike visual cameras that react to visible light spectrum, thermal sensors capture infrared radiations emitted by objects \cite{ITI}, which can bring to light the nuanced texture information about environmental surroundings even in challenging lighting conditions. 

Existing methods on RGB-T segmentation mainly focus on two aspects: cross-modal feature interaction between RGB and thermal images, and pixel-wise semantic analysis. To exploit cross-modality information from RGB-T pairs, early approaches adopt simple operations such as summation and concatenation \cite{mfnet, rtfnet, fuseseg, mffenet}. Recent attention-based fusion methods \cite{feanet, gmnet, ABMDRNet, mitigating, 10103760, li2022rgb, 9987529, zhou2022mtanet, Zhang_2023_CVPR} integrate RGB and thermal features by taking into consideration their relative context. However, channel-wise fusion directly overlaps RGB and thermal features, blind to their relative spatial positions (Figure~\ref{fig:intro} b.1). The `\textit{passive}' spatial-wise integration strategy directly provides features without asking for real needs (Figure~\ref{fig:intro} b.2). To achieve fine-grained fusion for semantic analysis, multi-supervision has been applied to semantically ambiguous regions~\cite{gmnet, mffenet}, but boundary supervision highly relies on accurate pixel-level labeling, which induces high-cost in practical applications. To better leverage the intrinsic contextual relativity between paired thermal signals and RGB features (e.g., to specifically and automatically target darkened areas in RGB images), a more target-guided proactive integration strategy is needed (Figure~\ref{fig:intro} b.3). 

In order to achieve this goal, we need to answer the following questions: $1)$ How to proactively use thermal signals to compensate the inadequate contextual semantics in optically-impaired regions? 
$2$) How to utilize a small set of RGB-T pairs with limited annotations to reach preferable segmentation performance for practical use?

To address the first challenge, we design a demand-guided target masking algorithm to enhance the features of poorly captured regions in RGB images in a `\textit{request}' manner, via proactive region-level masking with the learned compensation needs. To alleviate information loss incurred in encoding and maximize feature utility, we propose a spatial-aware recursive meshing method, which enhances cross-modal RGB-T features iteratively for pixel-wise semantic analysis. To fully exploit limited labeled pairwise data, we further propose an asymmetric data augmentation technique, named \textit{mono-modal CutOut} (M-CutOut), which creates artificial optically-impaired regions and encourages the network to learn more compensated features from thermal signals. An architecture design with semi-supervised learning capability is also introduced to utilize both natural and artificial regional complementarity in RGB-T. Extensive experiments on the popular MFNet and PST900 benchmarks demonstrate that our proposed framework, \textit{Spatial-aware Demand-guided Recursive Meshing} (SpiderMesh), achieves state-of-the-art performance on RGB-T semantic segmentation.

Our contributions are summarized as follows:
\begin{itemize}
  \setlength{\itemsep}{0pt}
  \setlength{\parsep}{0pt}
  \setlength{\parskip}{0pt}
  \item We propose a systematic framework, termed SpiderMesh, for RGB-T semantic segmentation. Specifically, a demand-guided target masking algorithm is proposed to directly meet the real needs of RGB-T feature compensation in a proactive `request’ manner, and a spatial-aware recursive meshing method to iteratively refine multimodal semantic features.
  \item To fully leverage the limited labeled pairwise data, we propose a data augmentation technique for RGB-T pairs and firstly extend the task to the semi-supervised setting.
  \item SpiderMesh not only achieves state-of-the-art performance, but also effectively addresses practical concerns such as computational complexity, robustness to signal loss, and manual labeling cost. 
\end{itemize}

\section{Related Work}

In this section, we briefly review two related topics, image semantic segmentation and RGB-T segmentation.

\subsection{Image Semantic Segmentation}
Image semantic segmentation is a pixel-level scene understanding task. Since FCN \cite{FCN} performed learning-based segmentation, early RGB methods \cite{segnet, unet} usually adopted the encoder-decoder network architecture. Deeplabv3 \cite{deeplabv3} proposed atrous spatial pyramid pooling (ASPP) to apply parallel atrous convolutions with different dilation rates, and SegFormer\cite{segformer} further boosted the performance. Although RGB segmentation methods have achieved promising progress in recent years, most methods are still susceptible to challenging lighting conditions with poor image quality. RGB-D (depth) methods \cite{fusenet, Wang_2018_ECCV, Cheng_2017_CVPR, Jiao_2019_CVPR, Xiong_2020_CVPR} leverage depth map to either enhance the whole RGB signal with depth value or highlight RGB features of foreground regions based on depth. However, RGB-D still falls short when additional semantic context for targeted areas is needed for strengthening poorly captured RGB regions.

\begin{figure*}
  \centering
  \includegraphics[scale=0.45]{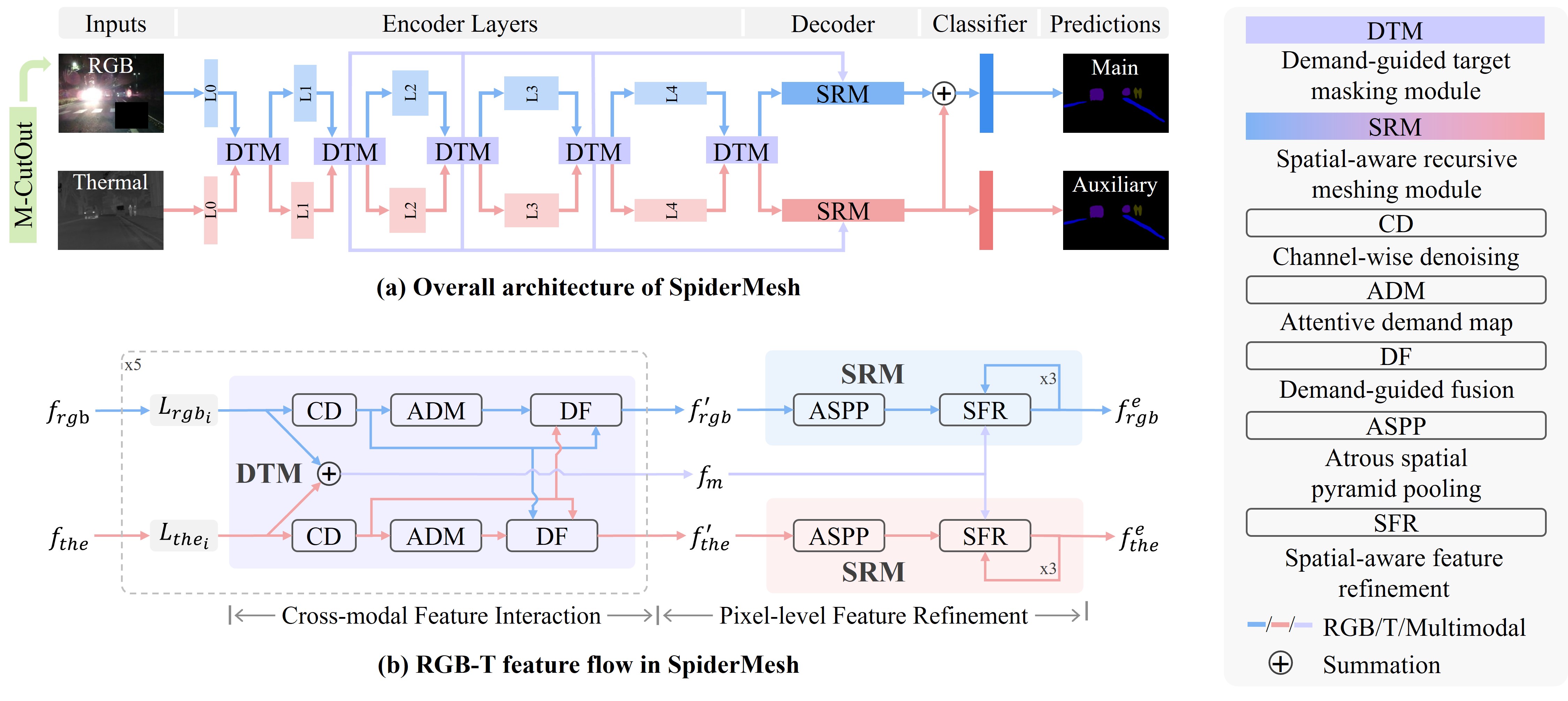}
  \caption{\textit{Top}: Overall architecture of SpiderMesh; \textit{Bottom}: RGB-T feature flow in SpiderMesh. $L_i$ denotes different layer of the backbone.}
  \label{fig:spidermesh}
\end{figure*}

\subsection{RGB-T Segmentation}
Thermal images can provide complementary information for those less informative regions in RGB images. Most of RGB-T fusion employed an explicit aggregation operation\cite{mitigating, 10103760, li2022rgb, 9987529, zhou2022mtanet, Zhang_2023_CVPR}. MFNet \cite{mfnet} collected an RGB-T semantic segmentation dataset and proved significantly performance improvement by utilizing thermal images. Two identical encoders were employed in RTFNet \cite{rtfnet} and FuseSeg \cite{fuseseg}, and the thermal features were gradually integrated into RGB features. FEANet \cite{feanet} refined the detail features using attention mechanism to deal with small objects. MFFENet \cite{mffenet} used spatial attention to emphasize foreground objects. The multimodal features are fused coarsely with simple operations (e.g., summation, concatenation) in these approaches. GMNet \cite{gmnet} proposed different fusion strategies for shallow and deep features to integrate multi-level features. Some recent works explored to utilize the power of transformer. MFTNet \cite{MFTNet} used modified transformer to learn intraspectral correlations and interspectral interaction, but introduced additional computational complexity. To further boost the performance, alignment-based fusion is utilized via domain adaptation techniques \cite{ABMDRNet, msuda}. Different from existing methods, we propose a systematic demand-guided approach addressing not only performance but also practical concerns. We enhance the less informative regions in RGB images with thermal features via proactive spatial-wise interaction. Instead of applying multi-supervision \cite{gmnet, mffenet}, we only utilize semantic supervision considering labeling cost, and enable the extension of our framework to semi-supervised semantic segmentation.

\section{SpiderMesh Framework }

In this section, we describe the proposed SpiderMesh framework, which consists of demand-guided target masking, spatial-aware recursive meshing, a novel M-CutOut technique for data augmentation, and a mutual learning strategy for semi-supervised adaptation.

\subsection{Overall Architecture}

As illustrated in Figure~\ref{fig:spidermesh}, RGB-T semantic segmentation is technically resolved into cross-modal feature interaction and pixel-level feature refinement in SpiderMesh framework. RGB-T pairs are fed into corresponding branches for each modality. Each branch adopts an encoder-decoder structure and employs ResNet \cite{resnet} as backbone for feature extraction. The number of input channels in the first convolutional layer of the thermal branch is set to $1$. Five encoder layers are utilized consecutively to extract features. A DTM (demand-guided target masking) module is embedded after each layer. Data scale is gradually decreased from $H \times W$ to $\frac{H}{32} \times \frac{W}{32}$. Next, a SRM (spatial-aware recursive meshing) module is used as the decoder to enhance unsampled features with fine-grained multimodal semantic features. We utilize bi-linear interpolation for upsampling. Although the two branches are treated equally during encoding and decoding, we regard the RGB branch as the main branch for generating final predictions. Thus, the enhanced thermal feature $f_{the}^{e}$ is introduced to the RGB branch and added with enhanced RGB feature $f_{rgb}^{e}$, which is further fed to a classifier. Meanwhile, $f_{the}^{e}$ is also fed to a classifier to output an auxiliary prediction. A Convolutional layer is used as the classifier.

\subsection{Demand-guided Target Masking}
\begin{figure}[ht]
  \centering
  \includegraphics[scale=0.35]{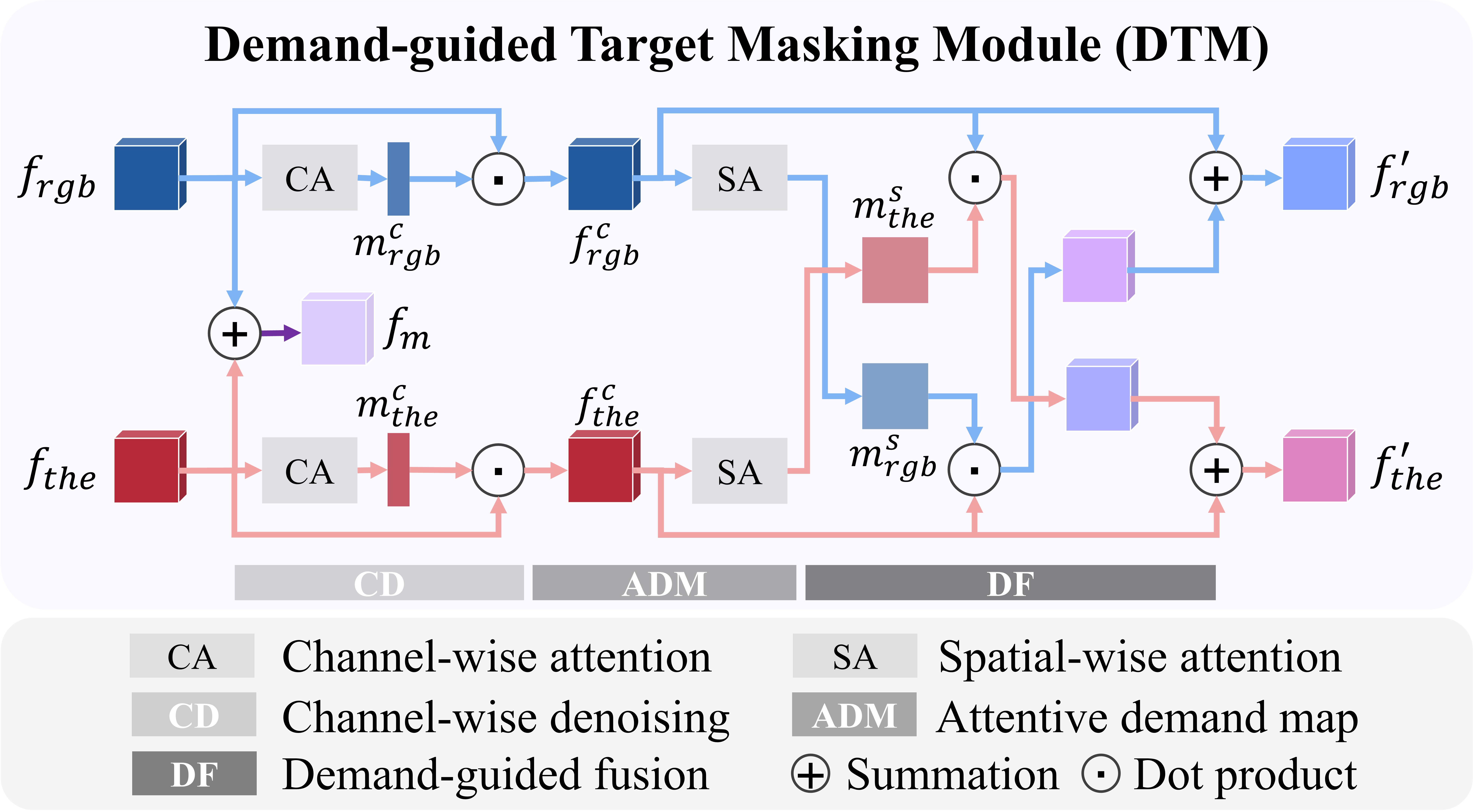}
  \caption{Architecture of DTM module. The less informative regions are complemented via demand-guided target masking in a `request' manner.}
  \label{fig:dtm}
\end{figure}

To better leverage the regional complementary texture feature across RGB and thermal signals, we propose a DTM module, a proactive spatial-wise fusion component whose architecture is illustrated in Figure~\ref{fig:dtm}. The overall feature interaction is guided by the demand map dynamically learned via spatial-wise attention. Technically, there are various implementation options for attention-based operations. We adopt a statistical approach \cite{cbam} as an example to demonstrate our insight of proactive spatial-wise fusion.

Take the RGB modality as an example, demand-guided target masking includes the following steps:

\textbf{Channel-wise Denoising.} To eliminate the inevitable camera noise (e.g., over-exposure lighting) before the demand map generation, $f_{rgb} \in \mathbb{R}^{H \times W \times C}$ is first denoised via channel-wise attention. The attention map $m_{rgb}^c \in \mathbb{R}^{1 \times 1 \times C}$ is used to weigh $f_{rgb}$ by element-wise multiplication, resulting in $f_{rgb}^c \in \mathbb{R}^{H \times W \times C}$.
\begin{equation}
  f_{rgb}^c = m_{rgb}^c \cdot f_{rgb} = \mathit{CA}(f_{rgb}) \cdot f_{rgb}
\end{equation}
where $\mathit{CA}(\cdot)$ is channel-wise attention operator.

\textbf{Attentive Demand Map.} To generate spatial-wise demand for thermal signal complementation, max-pooling and mean-pooling are utilized for spatial-wise statistics. The pooled features are concatenated and forwarded to a convolution operation with a filter in the size of $7 \times 7$ for further region-level statistics. After a sigmoid operation, the attentive demand map $m_{rgb}^s \in \mathbb{R}^{H \times W \times 1}$ is obtained, representing the demand of spatial-wise complementation for $f_{rgb}^c$.

\textbf{Demand-guided Fusion.} The thermal feature $f_{the}^c$ is spatial-wise weighted according to the adaptive demand represented by $m_{rgb}^s$. Then, $f_{rgb}^c$ is integrated with $f_{the}^c$ attentively in a `request' manner:
\begin{equation}
  \begin{aligned}
    f_{rgb}^{'} &= f_{rgb}^c + m_{rgb}^s \cdot f_{the}^c \\
             &= f_{rgb}^c + \mathit{SA}(f_{rgb}^c) \cdot f_{the}^c
  \end{aligned}
\end{equation}
where $\mathit{SA}(\cdot)$ is spatial-wise attention operator.

For the thermal modality, $f_{the}^{'}$ can be obtained likewise. In addition, input features $f_{rgb}$ and $f_{the}$ are also fused using summation operation to generate multimodal feature $f_{m}$ for detailed semantic feature refinement in later stage.

\subsection{Spatial-aware Recursive Meshing}
\begin{figure}[htbp]
  \centering 
  \includegraphics[scale=0.35]{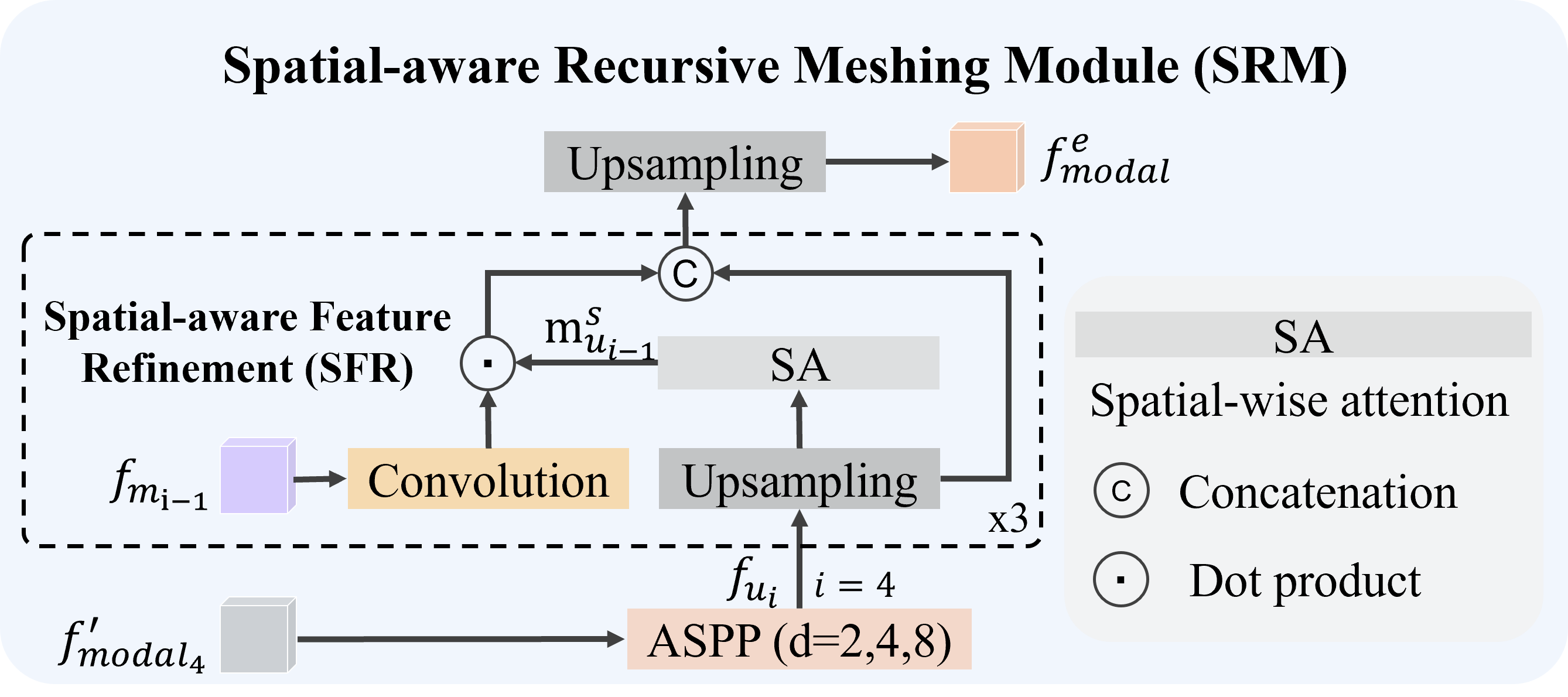}
  \caption{Architecture of SRM module. \textit{`modal'} is replaced with `rgb' or `the' according to which branch it applied in.}
  \label{fig:srm}
\end{figure}

Semantic segmentation is a pixel-level scene understanding task, which relies on fine-grained semantic features for pixel-wise classification. However, detailed information loss caused by downsampling is inevitable during encoding. To compensate information loss and refine the fine-grained features, we propose a SRM module that leverages the fused multimodal features in a recursive manner with spatial awareness. SRM module is composed of a modified ASPP block \cite{deeplabv3} and three spatial-aware feature refinement blocks, as shown in Figure~\ref{fig:srm}.

For RGB modality, the encoded feature map $f_{rgb_4}^{'}$ is first embedded with more global features via atrous spatial pyramid pooling. We use three dilation rates ($d=2,4,8$), and the number of channels is reduced to 256 after this block. To refine the fine-grained semantic features, the features of different receptive fields are recursively introduced via skip-connection. Instead of simple concatenation of features, we perform spatial-aware feature refinement to proactively mesh upsampled features with multimodal features $f_{m_i}$ which contain rich detailed semantic information from both RGB-T signals. Considering the complexity, channel reduction is applied to $f_{m_i}$ via convolutional operation. For features with a scale index $i$, the refinement step can be formulated as:
\begin{equation}
  \begin{aligned}
    f_{u_{i-1}} &= E(f_{u_i}, f_{m_{i-1}}) \\
                &= Up(f_{u_i}) \oplus (\mathit{Conv}(f_{m_{i-1}}) \cdot m_{u_{i-1}}^s) \\
                &= Up(f_{u_i}) \oplus (\mathit{Conv}(f_{m_{i-1}}) \cdot \mathit{SA}(Up(f_{u_i})))
  \end{aligned}
\end{equation}
where `$\oplus$' is the concatenation operator, `$\mathit{SA}(\cdot)$' is the spatial-wise attention operator, and `$Up(\cdot)$' and `$Conv(\cdot)$' denote the upsampling and convolution operators, respectively. $m_{u_i}^s$ is the spatial-aware attentive mask generated via spatial-wise attention operation, which indicates where and how much the feature needs to be compensated. 

The input RGB feature is compensated recursively, which can be represented by:
\begin{equation}
  f_{rgb}^{e} = Up(E(E(E(ASPP(f_{rgb_4}^{'}), f_{m_3}), f_{m_2}), f_{m_1}))
\end{equation}
where `$ASPP(\cdot)$' is the atrous spatial pyramid pooling. Similarly, the encoded thermal feature is compensated recursively to generate $f_{the}^{e}$.

\subsection{M-CutOut Augmentation}
\begin{figure}[ht]
  \centering
  \includegraphics[scale=0.35]{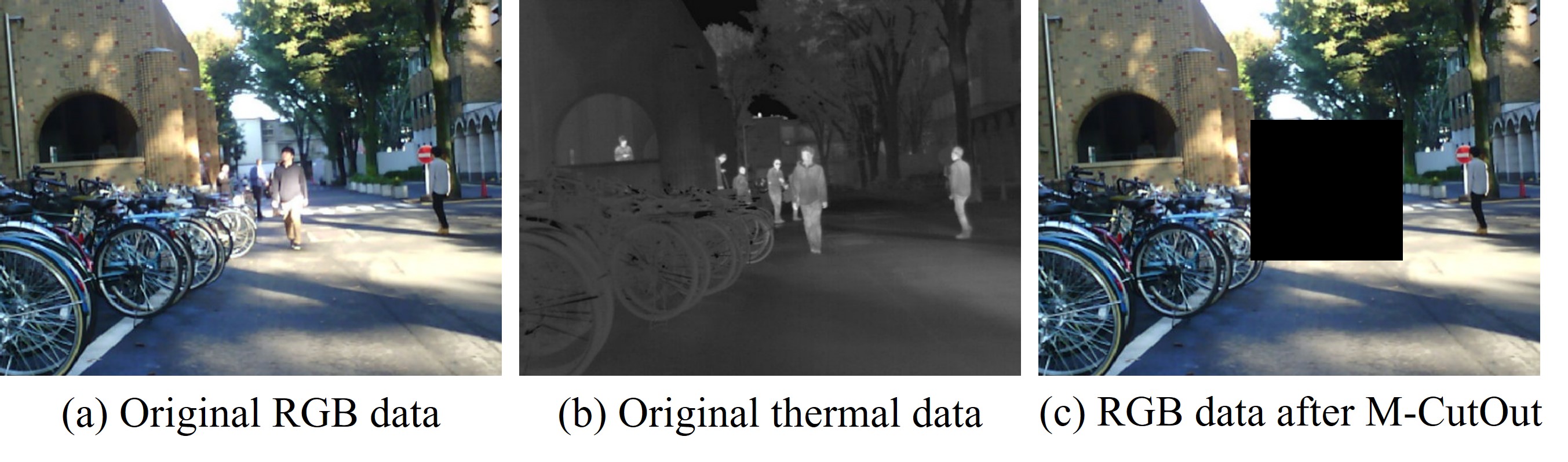}
  \caption{An example of M-CutOut. An optically-impaired region is created.}
  \label{fig:m-cutout}
\end{figure}

The key to RGB-T segmentation is to fully exploit the regional complementarity of thermal signals on optically-invisible regions. The Model is supposed to learn the intrinsic contextual relativity between RGB and thermal signals. However, normal CutOut \cite{cutout} masks all modalities. Different from that, M-CutOut cuts out part of the RGB image with randomly positioned mask $M$ and encourages the model to recover the masked RGB information with thermal signals, which is more in line with the nature of the task. To facilitate the learning of the adaptive cross-modal regional compensation, the artificial optically-impaired regions are randomly created. An example is shown in Figure~\ref{fig:m-cutout}.

\subsection{Semi-supervised Learning}

\begin{figure}[htbp]
  \centering
  \includegraphics[scale=0.37]{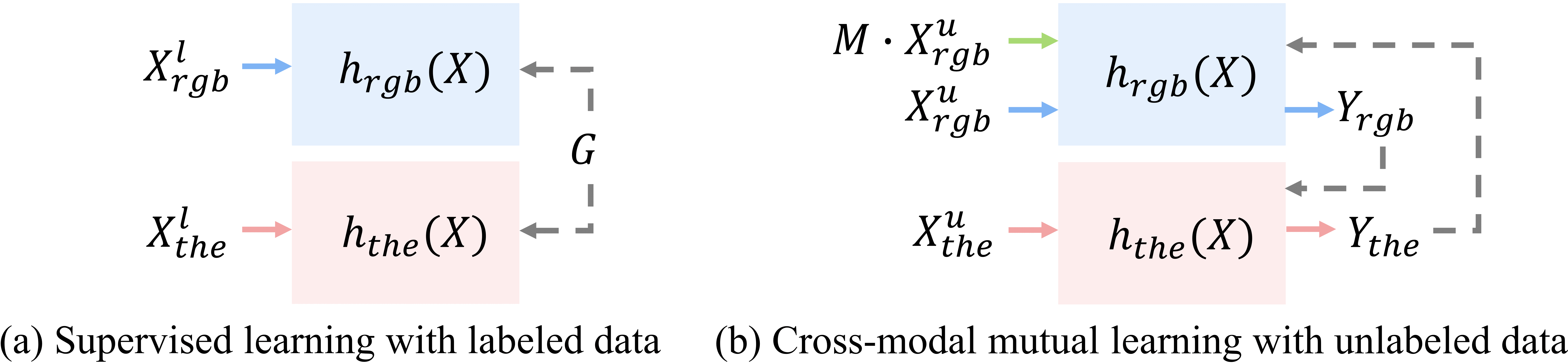}
  \caption{Framework for semi-supervised semantic segmentation.}
  \label{fig:semi}
\end{figure}

Benefiting from dual-branch architecture and M-CutOut, SpiderMesh can be easily extended to semi-supervised segmentation task by leveraging both natural and artificial regional complementarity in RGB-T (Figure~\ref{fig:semi}). 

Both labeled and unlabeled data are provided in semi-supervised learning tasks. Without loss of generality, let $G$ denote the label for labeled data pairs $(X_{rgb}^{l}, X_{the}^{l})$ and $(X_{rgb}^{u}, X_{the}^{u})$ denote unlabeled data. 
When using labeled data, the network is trained in a supervised manner:
\begin{equation}
  \mathcal{L_S} = \CE(G, h_{rgb}(X_{rgb}^{l})) + \CE(G, h_{the}(X_{the}^{l}))
\end{equation}
where `$h_{rgb}(\cdot)$' and `$h_{the}(\cdot)$' denote the two branches in SpiderMesh, and `$\CE(\cdot)$' is the cross-entropy loss function. Inspired by the applications of pseudo supervision in RGB segmentation task \cite{Chen_2021_CVPR, cpcl}, the network is trained via cross-modal mutual learning with unlabeled data for multimodal alignment-based fusion. 
We first generate the pseudo label $Y_{rgb}$ and $Y_{the}$ using predictions of original input data with weak augmentations only:
\begin{equation}
  Y_{rgb} = \argmax_y h_{rgb}(y|X_{rgb}^{u})
\end{equation}
$Y_{the}$ can be obtained likewise.
Then, we apply cross-modal pseudo supervision between the generated pseudo label and the predictions of augmented data using M-CutOut. The cross-modal supervision is conducive to modality alignment: 
\begin{equation}
  \mathcal{L_U} = \CE(Y_{the}, h_{rgb}(M \cdot X_{rgb}^{u})) + \CE(Y_{rgb}, h_{the}(X_{the}^{u}))
\end{equation}

The losses for supervised and unsupervised training are combined to form the final training objective:
\begin{equation}
  \mathcal{L} = \mathcal{L_S} + \mathcal{L_U}
\end{equation}

\begin{table*}
  \footnotesize
  \centering
  \caption{Quantitative evaluation on MFNet Dataset. Results of RGB and RGB-D semantic segmentation methods are obtained from \cite{rtfnet,mffenet,MFTNet,ABMDRNet}. The best values are marked by bold, and the second are marked by underline. All scores are in $\%$.}
  \begin{tabular}{cc|c|cccccccc} 
  \hline
        \textbf{Category}       &\textbf{Methods}                       & \textbf{mIoU}         & \textbf{Car}          & \textbf{Person}       & \textbf{Bike}         & \textbf{Curve}        & \textbf{Car-stop}     & \textbf{Guardrail}       & \textbf{Color-cone}      & \textbf{Bump}               \\ \hline
        
        \multirow{3}*{RGB}      & UNet\cite{unet}           & 45.1         & 66.2         & 60.5         & 46.2         & 41.6         & 17.9         & 1.8             & 30.6            & 44.2               \\
                                & SwinT\cite{swin}          & 49.0         & 85.2         & 57.6         & 61.0         & 33.2         & 28.0         & 2.4             & 42.7            & 33.5               \\
                                & BiSeNet\cite{bisenet}     & 50.0         & 84.1         & 63.2         & 60.1         & 36.7         & 25.3         & 5.0             & 42.2            & 35.9               \\ \hline
                            
        \multirow{2}*{RGB-D}    & SA-Gate\cite{Sagate}      & 45.8         & 73.8         & 59.2         & 51.3         & 38.4         & 19.3         & 0.0             & 24.5            & 48.8                \\
                                & ACNet\cite{acnet}         & 46.3         & 79.4         & 64.7         & 52.7         & 32.9         & 28.4         & 0.8             & 16.9            & 44.4                \\ \hline
                                
        \multirow{7}*{RGB-T}   & FuseSeg\cite{fuseseg}         & 54.5             & 87.9 & 71.7             & \underline{64.6}    & 44.8             & 22.7             & 6.4             & 46.9              & 47.9               \\
                                & ABMDRNet\cite{ABMDRNet}       & 54.8             & 84.8             & 69.6             & 60.3             & 45.1             & \underline{33.1} & 5.1             & 47.4              & 50.0               \\
                                & FEANet\cite{feanet}           & 55.3             & 87.8             & 71.1             & 61.1             & 46.5             & 22.1             & 6.6             & 55.3              & 48.9               \\
                                & MFFENet-single\cite{mffenet}  & 55.5             & 87.1             & \underline{74.4}    & 61.3             & 45.6             & 30.6             & 5.2             & \textbf{57.0}     & 40.5               \\
                                & MFTNet\cite{MFTNet}           & 57.3 & 87.9 & 66.8             & 64.4 & 47.1 & \textbf{36.1}    & \underline{8.4} & \underline{55.5}  & 62.2               \\
                                & DooDLENet\cite{DooDLeNet}     & 57.3 & 86.7             & 72.2             & 62.5             & 46.7             & 28.0             & 5.1             & 50.7              & \textbf{65.8} \\
                                
        \cline{2-11}            & SpiderMesh-152 (Ours)                   & \underline{57.9}    & \underline{88.1}    & 72.8 & 63.7             & \underline{48.4}    & 28.2             & \textbf{8.8}    & 48.2              & \underline{64.2}      \\ 
                                & SpiderMesh-B4 (Ours)                   & \textbf{58.4}    & \textbf{89.9}    & \textbf{75.3} & \textbf{64.8}             & \textbf{51.5}    & 31.4             & 4.5    & 54.5              & 55.9      \\\hline

  \end{tabular}
  \label{tab:sota_mf}
\end{table*}

\begin{table*}
  \footnotesize
  \centering
  \caption{Quantitative evaluation on PST900 Dataset. Results of compared baselines are obtained from \cite{mffenet}. The best values are marked by bold, and the second are marked by underline. All scores are in $\%$.}
  \begin{tabular}{cc|c|ccccc} 
  \hline
    \textbf{Category}       & \textbf{Methods}                       & \textbf{mIoU}         & \textbf{Survivor}     & \textbf{Hand-drill}   & \textbf{Backpack}     & \textbf{Fire-extinguisher}  & \textbf{Background}  \\ \hline

    \multirow{1}*{RGB}      & UNet\cite{unet}           & 52.8         & 31.6         & 38.3         & 52.9         & 43.0               & 98.0       \\ \hline
    \multirow{1}*{RGB-D}      & ACNet\cite{acnet}           & 71.8         & \underline{65.2}         & 51.5         & \underline{83.3}         & 60.0               & \underline{99.3}       \\ \hline
    \multirow{5}*{RGB-T}    & RTFNet\cite{rtfnet}           & 57.6         & 36.4         & 25.4         & 75.3         & 52.0               & 98.9        \\
                            & PSTNet\cite{pst900}           & 68.4         & 50.0         & 53.6         & 69.2         & 70.1               & 98.9        \\
                            & ABMDRNet\cite{ABMDRNet}           & 71.3         & 62.0         & 61.5         & 67.9         & 66.2               & 99.0        \\
                            & MFFENet-single\cite{mffenet}  & \underline{77.1} & 63.0 & \underline{66.8} & 76.6  & \textbf{79.8} & \underline{99.3} \\

    \cline{2-8}             & SpiderMesh-152 (Ours)                & \textbf{82.3} & \textbf{71.9} & \textbf{79.7} & \textbf{84.0} & \underline{76.6} & \textbf{99.4} \\ \hline

  \end{tabular}
  \label{tab:sota_pst}
\end{table*}

\section{Experiments}

In this section, we compare our model with other methods and provide detailed ablation studies on standard RGB-T semantic segmentation datasets.

\subsection{Experimental Setting}

\textbf{Datasets}
We evaluate the proposed SpiderMesh on public datasets of both urban scenes (MFNet \cite{mfnet}) and underground scenes (PST900 \cite{pst900}). 
MFNet Dataset \cite{mfnet} is the only public dataset on RGB-T semantic segmentation for urban traffic scenes. It contains $1,569$ pairs of RGB and thermal images captured simultaneously, which comprises 820 daytime and 749 nighttime paired images. For fair comparison, we follow the same splitting scheme as in previous work. Batch size is $6$, and the input is resized to a fixed size of $480 \times 640$.

PST900 is a challenging underground environment dataset proposed for the DARPA Subterranean Challenge \cite{pst900}. It contains $894$ aligned RGB-T pairs collected from diverse environments with varying lighting conditions. We adopt the same splitting scheme as in \cite{pst900} for fair comparison. Input data is resized to $720 \times 1280$, and batch size is $2$.


\textbf{Implementation Details}
We mainly employ ResNet-152 as backbone. The encoder is initialized with the pre-trained weights provided by PyTorch. The initial learning rate is set to $10^{-2}$, and exponential decay scheme is adopted to gradually decrease the learning rate. We use SGD optimizer with momentum for training. The momentum and weight decay are set as $0.9$ and $5 \times 10^{-4}$. The network is trained until convergence (200 epochs). For training, we apply several data augmentation methods, including random flipping, random cropping, and the proposed M-CutOut. Experiments are implemented in PyTorch on a server with NVIDIA A30.

\subsection{Main Results}

Table~\ref{tab:sota_mf} reports the comparison results on MFNet dataset. Besides ResNet-152, we also report the performance with MiT-B4 as a reference point for transformer-based approach. SpiderMesh-B4 achieves the best performance on mIoU ($58.4\%$), and outperforms baselines on $4$ categories (car, person, bike, and curve). Among them, cars, pedestrians and bikes are the three most common objects in urban scenes. SpiderMesh-152 is $0.5\%$ lower than SpiderMesh-B4, with the best IoU on guardrail and the second best on $3$ categories (car, curve, and bump). Most of RGB-T methods outperforms both RGB and RGB-D techniques, which demonstrates the importance of tackling RGB-T segmentation in a task-specific way to leverage the regional complementarity. The reported performance of MFFENet-single \cite{mffenet} is only under semantic supervision for fair comparison. Although the full-version GMNet can achieve 57.3 \% mIoU utilizing multi-supervision, its performance drops to 53.9\% when boundary supervision is not applied \cite{gmnet}. There are performance fluctuations on the \textit{guardrail} category since the unbalanced class distribution (the proportion of the \textit{guardrail} class is $0.095\%$ \cite{mffenet}). Although SpiderMesh-B4 performs better, the complexity is also higher than the ResNet-152 version, so we choose SpiderMesh-152 for the remaining experiments considering practical application.



To further evaluate the proposed SpiderMesh, we also compare it with baselines on PST900 dataset, as reported in Table~\ref{tab:sota_pst}. In line with expectations, SpiderMesh consistently outperforms others on mIoU under diverse underground scenes. For $4$ foreground categories, it achieves the best performance on $3$ (survivor, hand-drill, and backpack). SpiderMesh achieves a performance increase of $8.9\%$ on \textit{survivor} class over MFFENet-single, by effectively leveraging regional complementary features from thermal images.

\subsection{Ablation Study}

To better understand SpiderMesh, we conduct several groups of experiments for ablation study on MFNet dataset.

\textbf{Effect of Each Component.} In the baseline network, multimodal features are only fused at the classifier in the RGB branch, and normal feature upsampling operations are adopted. As we can see from Table~\ref{tab:ablation}, the overall gain of the three components is $5.5\%$. Among them, performance improvement from DTM and M-CutOut are more significant, thanks to their regional complementation for RGB regions with thermal images. SRM further compensates the spatial-wise information loss with fused multimodal features and leads to an improvement of $0.7\%$.

\begin{table}[ht]
  \footnotesize
  \centering
  \caption{Ablation study on SpiderMesh.}
  \begin{tabular}{cl|c} 
  \hline
    \multicolumn{2}{c|}{\textbf{Ablated SpiderMesh}}                 & \textbf{mIoU (\%)} \\ \hline
    - & \cellcolor{lightgray} \textbf{Baseline}  & \cellcolor{lightgray} \bf52.4 \\ \hline
    \multirow{3}*{+ Fusion}     & Summation during encoding       & 53.4 \\ 
                                & Cross-modal weighted fusion     & 54.5 \\
        & \cellcolor{lightgray} \textbf{DTM} & \cellcolor{lightgray} \bf55.2 \\ \hline
    \multirow{3}*{+ Refinement} & SRM w/ self-modal feature only  & 55.4 \\
                                & SRM w/ cross-modal feature only & 55.4 \\
        & \cellcolor{lightgray} \textbf{SRM} & \cellcolor{lightgray} \bf55.9 \\ \hline
    \multirow{2}*{+ Data aug.} & normal CutOut            & 56.0 \\
        & \cellcolor{lightgray} \textbf{M-CutOut} & \cellcolor{lightgray} \bf57.9 \\ \hline

  \end{tabular}
  \label{tab:ablation}
\end{table}

\begin{figure}[ht]
  \centering
  \includegraphics[scale=0.3]{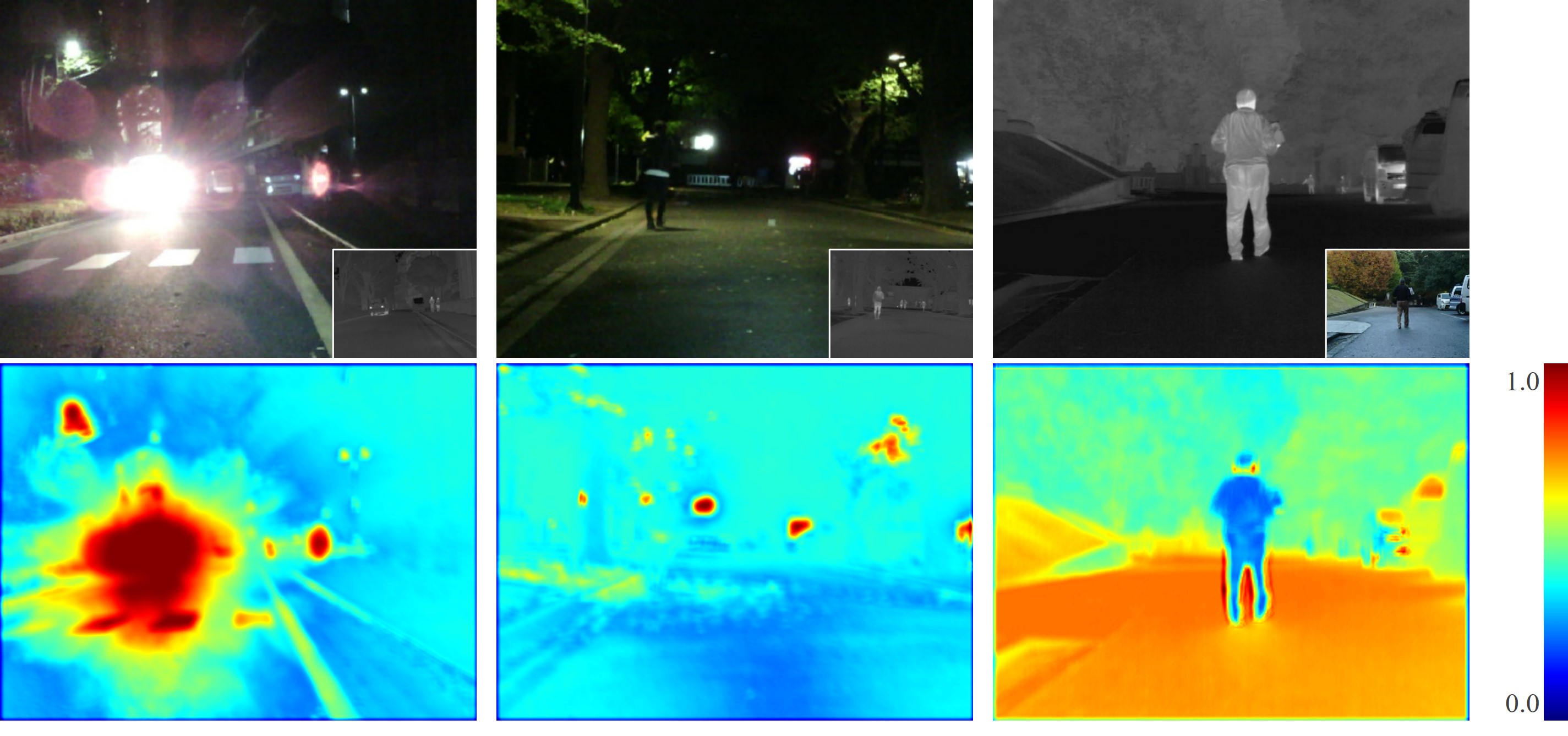}
  \caption{Visualization of demand map for requesting cross-modal regional complementary information. \textit{Left} and \textit{Middle} are RGB demand maps in dazzling-light and nighttime. \textit{Right} is the thermal demand map in daytime.}
  \label{fig:dmap}
\end{figure}

\textbf{Architecture Design.} Firstly, we explore different RGB-T fusion methods in DTM. The simple feature summation during encoding can improve mIoU to $53.4\%$, but the features are fused without distinction. The cross-modal weighted fusion is in a passive `post’ manner, where one modality is spatial-wise weighted via self-attention and then integrated to the other modality. It performs better than summation, but is still not as good as the proactive manner in DTM. Figure~\ref{fig:dmap} shows that the less informative an area is, the higher its demand for fusion is. For example, the dark and overexposed areas of RGB images usually have higher compensation demands, corroborating our hypothesis on the regional complementary power of thermal signals. Secondly, we study the design choices for SRM, and the refinement with fused multimodal feature is a better choice since it contains rich detailed semantic information from both RGB-T signals. Besides, the benefit of normal CutOut is limited, and the replacement to M-CutOut results in a lift of $1.9\%$.

\textbf{Robustness Analysis.} It is important to analyze the robustness of multimodal approaches, as signal loss due to software/hardware failure is common in practice. To test SpiderMesh, we manually set the input of a modality to $0$ to simulate the signal loss, and evaluate the performance in daytime and nighttime scenarios. The results of both branches are reported in Table~\ref{tab:robust} for analysis. Overall, the performance with inputs from both modalities is the best, which demonstrates the benefit of leveraging multimodal information. The performance differences between RGB-only and thermal-only input indicate that thermal is the more dominant modality due to its high reliability under poor lighting conditions. The slight advantage of the RGB branch comes from further feature fusion at the classifier. As we can see, SpiderMesh can still generate valid predictions when facing input signal loss.

\begin{table}[t]
  \footnotesize
  \centering
  \caption{Robustness analysis on modality signal loss (\%).}
  \begin{tabular}{cc|ccc} 
  \hline

    \textbf{Input modality}                    & \textbf{Branch}     &  \textbf{Daytime}           & \textbf{Nighttime}       & \textbf{All time}      \\ \hline
    \multirow{2}*{RGB + thermal}               & RGB                 &  \textbf{52.0}              & \textbf{56.0}            & \textbf{57.9} \\
                                               & Thermal             &  51.0                       & 55.7                     & 57.3          \\ \hline
    \multirow{2}*{RGB only}                    & RGB                 &  \textbf{40.1}              & \textbf{32.5}            & \textbf{39.6} \\
                                               & Thermal             &  39.8                       & 32.4                     & 39.2          \\ \hline
    \multirow{2}*{Thermal only}                & RGB                 &  \textbf{41.7}              & \textbf{51.1}            & \textbf{50.5} \\
                                               & Thermal             &  41.6                       & 50.8                     & 50.2          \\ \hline
    
  \end{tabular}
  \label{tab:robust}
\end{table}

\textbf{Complexity Analysis} We obtain several versions of SpiderMesh by replacing the backbone. The complexity and corresponding performances are summarized in Table~\ref{tab:complexity}. SpiderMesh-B4 achieves the best performance with the highest complexity. Considering the computational cost in practice, SpiderMesh-152 achieves a better tradeoff and has advantages on both performance and complexity compared with the two representative methods also using ResNet-152. The model can be more lightweight by employing lighter backbone, and even SpiderMesh-50 ($54.4\%$) can be on par with other RGB-T approaches, like FuseSeg \cite{fuseseg} ($54.5\%$). 

\begin{table}[ht]
  \footnotesize
  \centering
  \caption{Complexity analysis on different versions of SpiderMesh.}
  \begin{tabular}{cc|cc} 
  \hline

    \textbf{Version}       & \textbf{Backbone}     &  \textbf{GFlops}      & \textbf{mIoU (\%)}   \\ \hline
    RTFNet\cite{rtfnet}     & ResNet-152   & 290.6       & 53.2        \\
    MFTNet\cite{MFTNet}     & ResNet-152   & 330.6        & 57.3        \\ \hline
    SpiderMesh-50    & ResNet-50    &  168.2       & 54.4        \\
    SpiderMesh-101   & ResNet-101   &  214.0       & 56.1        \\
    SpiderMesh-152   & ResNet-152   &  259.8       & 57.9        \\
    SpiderMesh-B4    & MiT-B4       &  398.8       & 58.4        \\ \hline
  \end{tabular}
  \label{tab:complexity}
\end{table}

\subsection{Evaluation on Semi-supervision}
We further evaluate SpiderMesh under semi-supervision setting. The 784 labeled images are randomly split into two subsets, 392 images are regarded as unlabeled subset. We make sure that each class appears in the labeled subset. Making use of unlabeled data, SpiderMesh yields a performance lift of \textbf{2.1\%} (from \textbf{53.2\%} to \textbf{55.3\%}). The performance in low-data regime is comparable with that of other methods under full supervision as reported in Table~\ref{tab:sota_mf}. Considering the randomness in splitting scheme, we conduct 5 experiments with different partitions, and the gain is $2.1 \pm 0.1\%$ leveraging unlabeled pairwise RGB-T data.

 


\section{Conclusion}
In this paper, we manage to fully exploit the regional complementarity of thermal signals on optically-invisible regions. The systematic SpiderMesh framework is proposed. DTM proactively compensates the features of less informative regions via demand-guided target masking in a `request’ manner, SRM recursively refines detailed semantic features for segmentation task. M-CutOut is proposed to create new optically-impaired regions and encourage the model to learn compensated features. Besides, a semi-supervised setting is first explored by leveraging both natural and artificial regional complementarity, which deserves further study.


\bibliographystyle{IEEEtran}
\bibliography{mybib}

\end{document}